\documentclass[final]{neus2025}

\usepackage{subcaption} 
\usepackage{times, pifont, hyperref, tcolorbox}
\usepackage{epigraph}
\usepackage{enumitem} 

\title[Video Agents Challenge]{A Challenge to Build Neuro-Symbolic Video Agents}

\coltauthor{
    \Name{Sahil Shah} \Email{ss96869@utexas.edu}\\
    \Name{Harsh Goel} \Email{harshg99@utexas.edu}\\
    \Name{Sai Shankar Narasimhan} \Email{nsaishankar@utexas.edu}\\
    \Name{Minkyu Choi} \Email{minkyu.choi@utexas.edu}\\
    \Name{S P Sharan} \Email{spsharan@utexas.edu}\\
    \Name{Oguzhan Akcin} \Email{oguzhnakcin@utexas.edu}\\
    \Name{Sandeep Chinchali} \Email{sandeepc@utexas.edu}\\
 \addr The University of Texas at Austin
 }

\usepackage{xparse}

\newcommand{\todo}[1]{\textcolor{yellow}{TODO: fix commas}}

\DeclareDocumentCommand{\query}{m o}{%
    \IfNoValueTF{#2}{%
        t_{#1} 
    }{%
        t_{#1,#2} 
    }%
}

\DeclareDocumentCommand{\spant}{m o}{%
    \IfNoValueTF{#2}{%
        y_{#1} 
    }{%
        y_{#1,#2} 
    }%
}

\makeatletter
\DeclareRobustCommand\onedot{\futurelet\@let@token\@onedot}
\def\@onedot{\ifx\@let@token.\else.\null\fi\xspace}

\def\eg{\emph{e.g}\onedot}

\begin{document}

\maketitle

\begin{abstract}
Modern video understanding systems excel at tasks such as scene classification, object detection, and short video retrieval. However, as video analysis becomes increasingly central to real-world applications, there is a growing need for \textit{proactive} video agents---systems that not only interpret video streams but also reason about events and take informed actions. A key obstacle in this direction is \textit{temporal reasoning}: while deep learning models have made remarkable progress in recognizing patterns within individual frames or short clips, they struggle to understand the sequencing and dependencies of events over time, which is critical for action-driven decision-making. Addressing this limitation demands moving beyond conventional deep learning approaches. We posit that tackling this challenge requires a \textit{neuro-symbolic} perspective, where video queries are decomposed into atomic events, structured into coherent sequences, and validated against temporal constraints. Such an approach can enhance interpretability, enable structured reasoning, and provide stronger guarantees on system behavior, all key properties for advancing trustworthy video agents. To this end, we present a grand challenge to the research community: developing the next generation of intelligent video agents that integrate three core capabilities---(1) autonomous video search and analysis, (2) seamless real-world interaction, and (3) advanced content generation. By addressing these pillars, we can transition from \textit{passive} perception to intelligent video agents that reason, predict, and act, pushing the boundaries of video understanding.
\end{abstract}


\section{Introduction}
\label{sec:intro}
\begin{figure}[t]
\captionsetup{type=figure}
    \centering
    \includegraphics[width=0.98\linewidth]{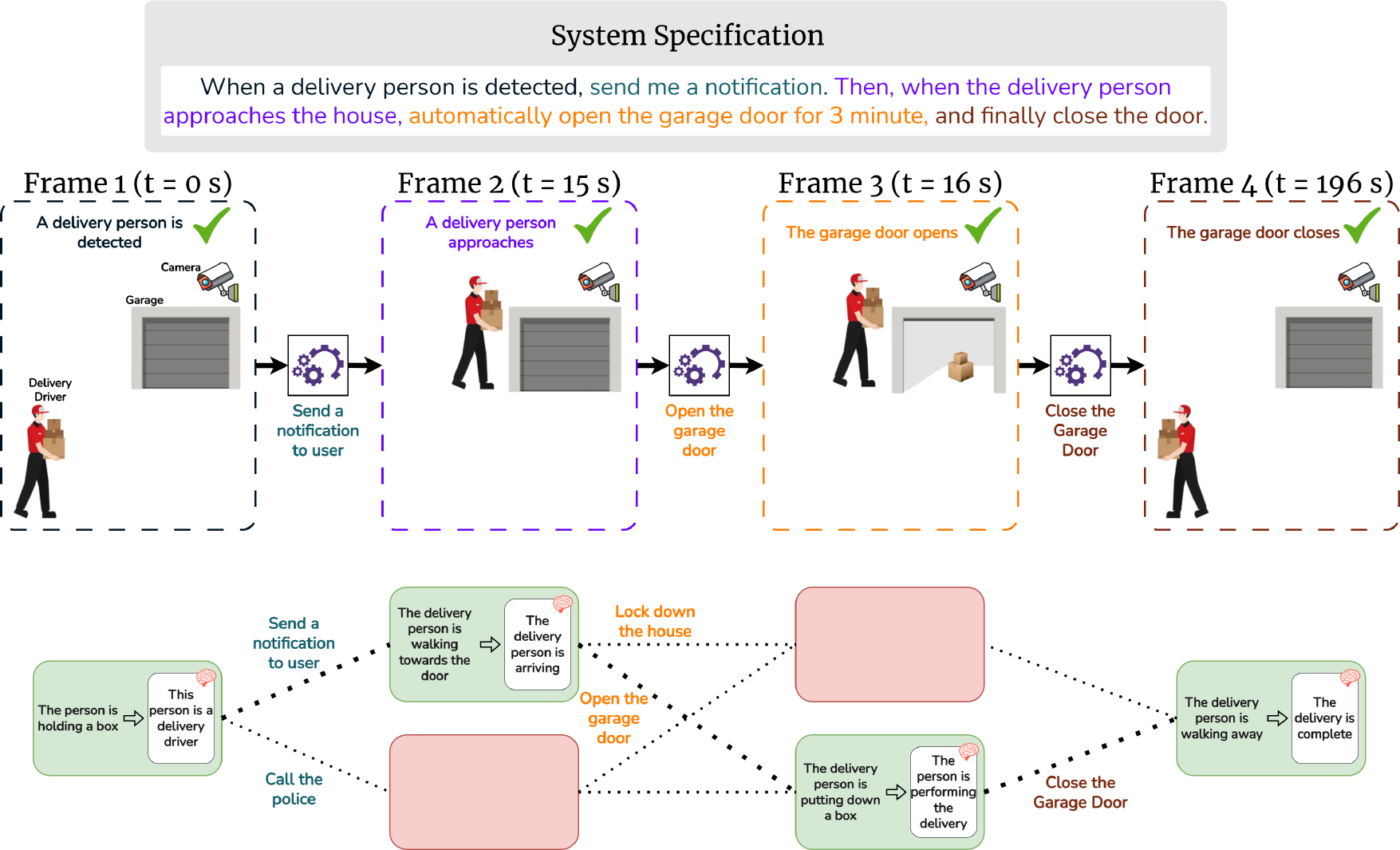}
    \captionof{figure}{\textbf{An Efficient Neuro-Symbolic Approach to Video Agents.} We argue for a neuro-symbolic approach to develop video agents that combines the per-frame or short-horizon reasoning capabilities of neural perception models with the long-term reasoning abilities of symbolic frameworks such as temporal logic tools. Here, we show one such example from a home security system, where the agent is required to identify the presence of a delivery person and send the required notifications.}
    \label{fig:teaser}
\end{figure}

Consider an agentic system that utilizes video feed from the cameras of a home security system (Figure \ref{fig:teaser}) to perform the following complex user request:

\begin{quote}
\textit{ Send me a notification when a person is walking up to my driveway with a package, then automatically open the garage for three minutes to safely store the package and lock everything back up when the delivery is complete.}
\end{quote}

Home-security owners increasingly expect video camera systems to handle tasks of this complexity \citep{comeau2024integrated}. Beyond delivery automation, such a system should also be able to notify law enforcement authorities if it recognizes an attempted break-in, send the exact video footage of the incident, and autonomously trigger an alarm to lock all entry points to secure the premises. Requests of this nature are not limited to home security; similar demands arise in defense \citep{shultz2020bigdata}, autonomous driving \citep{nhtsa2024aeb}, and social media analytics \citep{metricom2025trends}. Endowing current video systems to process user queries as shown above requires moving beyond \textit{passive} analysis to \textit{reason} about unfolding events in real-time, and precisely \textit{interact} with the real world with guarantees.

While deep learning excels at understanding short-term activities (\eg, detecting a person, object, or short event), it struggles with temporal reasoning and long-term memory (see Section \ref{sec:dlvideo}), posing a significant challenge for video agents tasked with understanding complex user queries \citep{choi2024towards}. Furthermore, deep learning methods in isolation do not provide the necessary interpretability or guarantees for either perceiving or acting. 

In contrast, structured logic representations, such as temporal logic (TL) \citep{Baier2008}, effectively address these challenges by explicitly encoding time-dependent sequences of events such as ``close the garage door within 20 minutes of opening it.'' Therefore, we argue that the next generation of video understanding systems will adopt a hybrid of deep learning (\textit{neuro}) and formal representations such as TL (\textit{symbolic}). A \textbf{neuro-symbolic} approach can capture complex temporal constraints while providing interpretability and formal guarantees with respect to user specifications. Therefore, we posit that future video systems---which we term \textbf{video agents}---will be built on three pillars:
\begin{enumerate}
    \item \textbf{Video Search and Understanding}: A video agent must identify and extract the relevant frames that correspond to a complex temporally extended query. For instance, in the running example, tagging a person with a package walking towards the front door as an attempted delivery would require the system to process the progression of events over time.

    \item \textbf{Integrate Understanding with Real-World Action:} Video agents must act in the real world according to user specifications by integrating perception with downstream actions such as calling relevant Application Programming Interfaces (APIs) \citep{langchain}. For example, actions such as ``Opening the garage for three minutes after detecting a delivery'' or ``Locking down entry points during an attempted break-in'' are enacted only after specific sequences of events are identified.

    \item \textbf{Video Generation:} Video agents must be rigorously tested through synthetic simulations for edge cases, for instance, distinguishing between a delivery driver and an intruder or detecting suspicious packages. Additionally, video generation serves as a tool for these agents to create video clips that edit out sensitive user content for post-hoc analysis, for instance, in our example, generating or editing video clips for law enforcement or delivery companies.
    
\end{enumerate}

Hence, we propose a grand vision and challenge for the community to develop the next generation of neuro-symbolic video agents capable of analyzing videos (both offline and in real time) and executing user-specified actions. We outline objectives for each desired capability---video search, action execution, and video generation---and provide a starting point with relevant datasets (TLV \citep{choi2024towards} for video search) and metrics (NeuS-V \citep{sharan2024neuro} for video generation) to motivate further research and development in this direction. Finally, we urge the research community to establish rigorously standardized benchmarks and systematically validate video agents as these agents transition into safety-critical domains such as autonomous driving and home security.

\section{Can Deep Learning Alone Solve Video Understanding?}
\label{sec:dlvideo}
\begin{figure}[t]
    \floatconts
    {fig:combined_methods}
    {\caption{\textbf{NSVS-TL and NeuS-V System Diagrams.} In (\textit{a}), when given video feed from a security system, NSVS-TL demonstrates its capability in identifying exactly when a delivery driver walks up the stairs and drops a package off. Similarly, in (\textit{b}), a video generated by a foundation model describing a delivery scenario is evaluated for temporal fidelity through NeuS-V.}}
    {
        \subfigure[NSVS-TL Architecture.]{
            \label{fig:nsvssample}
            \includegraphics[width=0.46\linewidth]{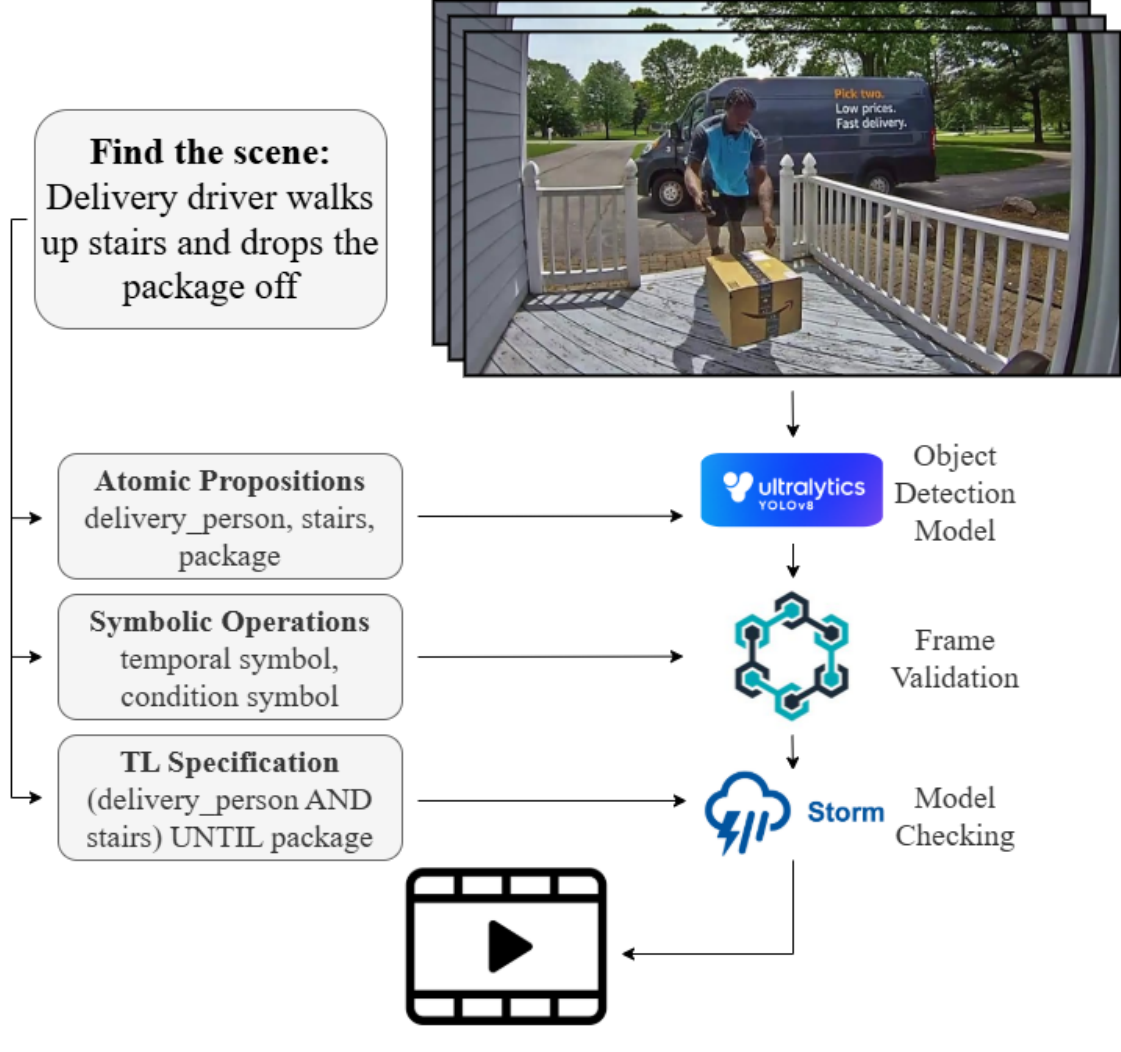}
        }
        \hfill
        \subfigure[NeuS-V Architecture.]{
            \label{fig:neusvsample}
            \includegraphics[width=0.46\linewidth]{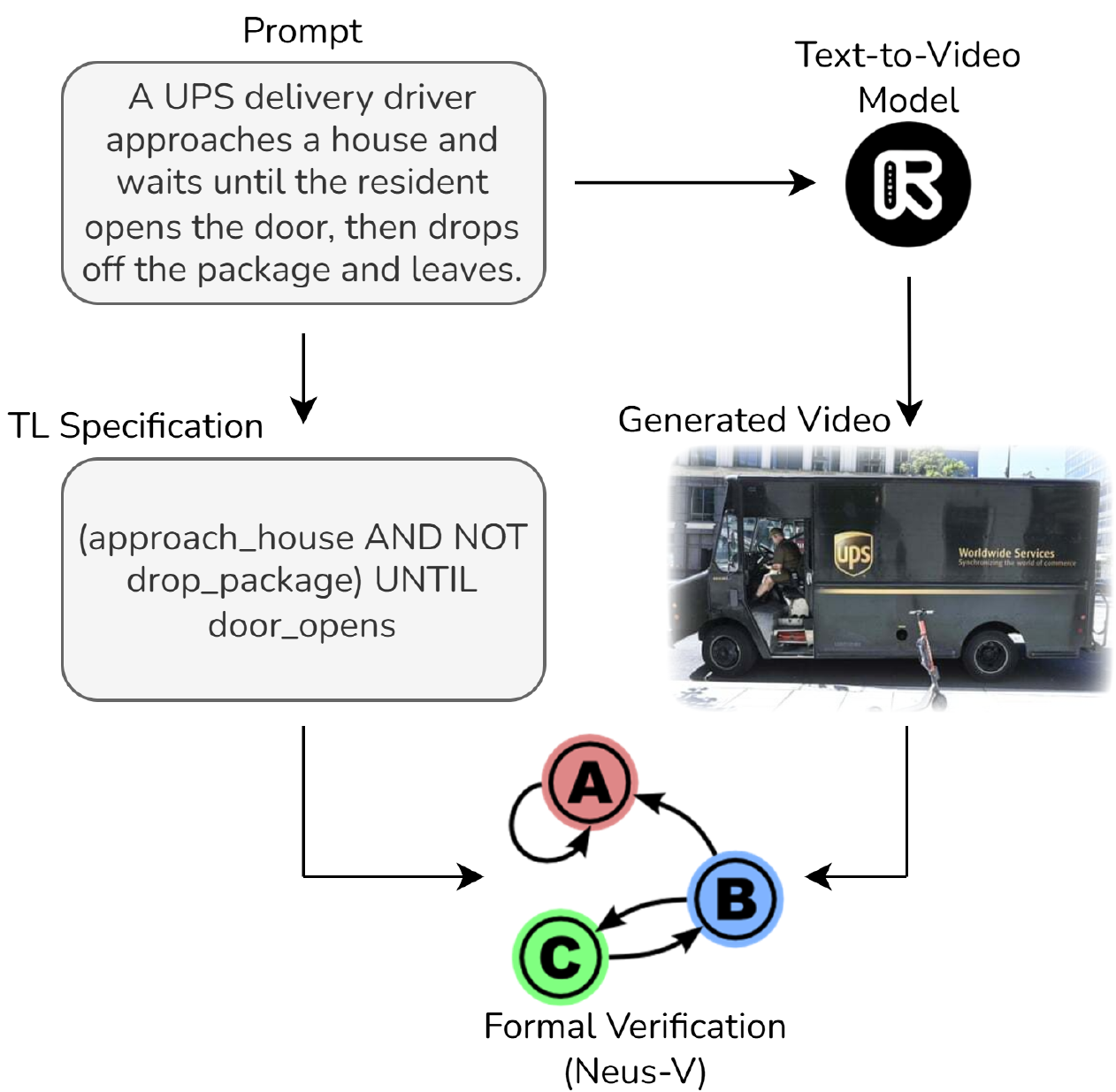}
        }
    }
\end{figure}

In this section, we revisit a fundamental question: Can deep learning alone achieve comprehensive video understanding, the key building block for any video agent? While state-of-the-art multimodal foundation models such as GPT4 \citep{achiam2023gpt}, LLaMA \citep{touvron2023llama}, and Gemini \citep{team2023gemini} have demonstrated impressive capabilities for language and image tasks, these models lack the ability to interpret extended temporal dependencies between events. In the following, we highlight the key drawbacks of these foundation models in the context of video search and generation, both critical for the agent's operation.

\begin{figure}[t]
    \floatconts
    {fig:combined_results} 
    {\caption{\textbf{Foundation models struggle to perform video search and generation with increasing complexity of user queries.} However, neuro-symbolic approaches (NSVS-TL) effectively decouple spatial and temporal reasoning using perception modules for spatial reasoning and temporal logic (TL) to model long-term temporal dependencies. As a result, NSVS-TL outperforms foundation models in complex video search tasks (\textit{a}). Similarly, text-to-video models, like Pika, fail to maintain temporal consistency as scenario complexity increases (\textit{b}).}}
    {
        \subfigure[NSVS-TL excels at complex video search.]{
            \label{fig:nsvstl}
            \includegraphics[width=0.46\linewidth]{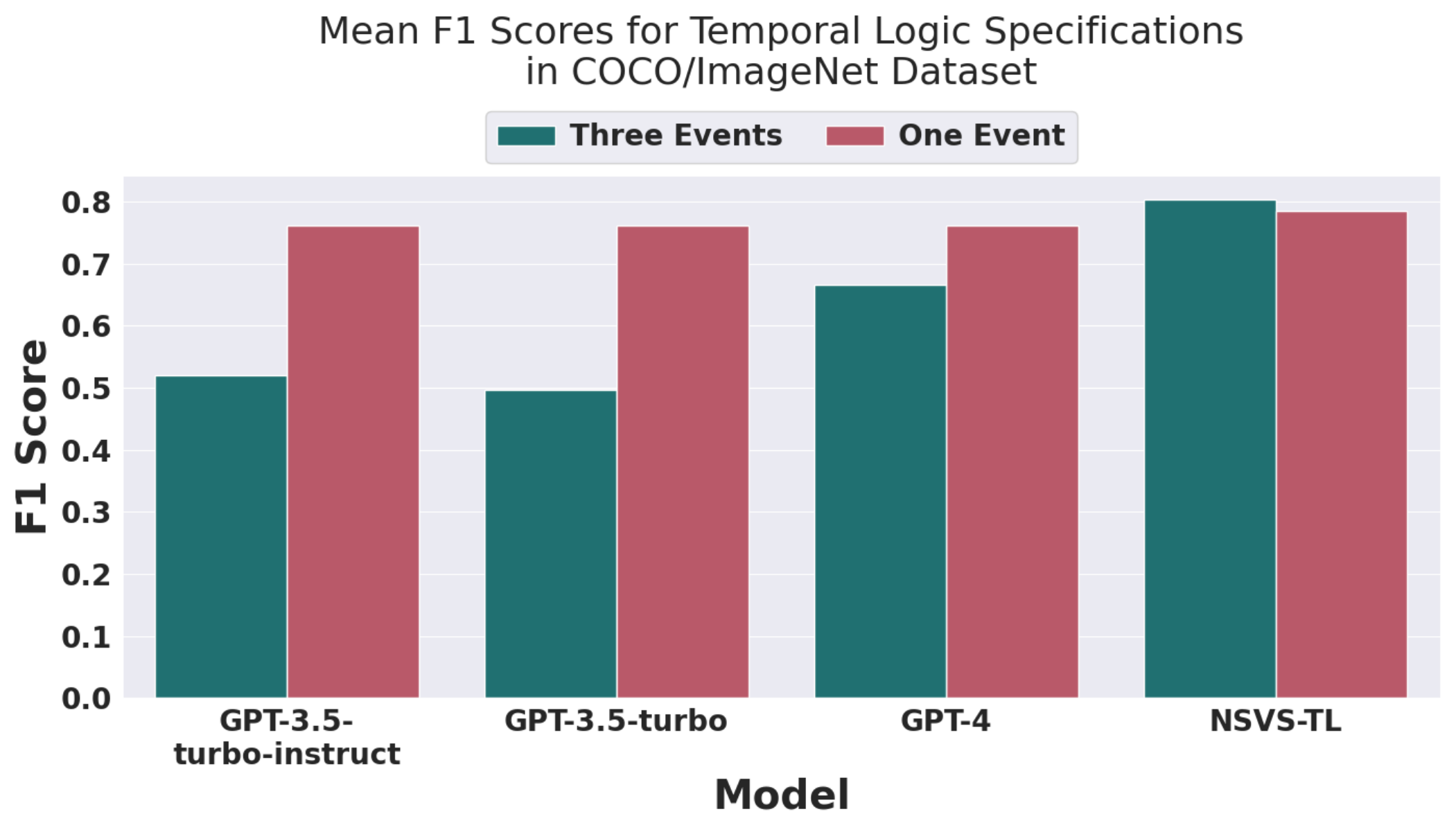}
        }
        \hfill
        \subfigure[Video generation fails with complexity.]{
            \label{fig:neusv}
            \includegraphics[width=0.46\linewidth]{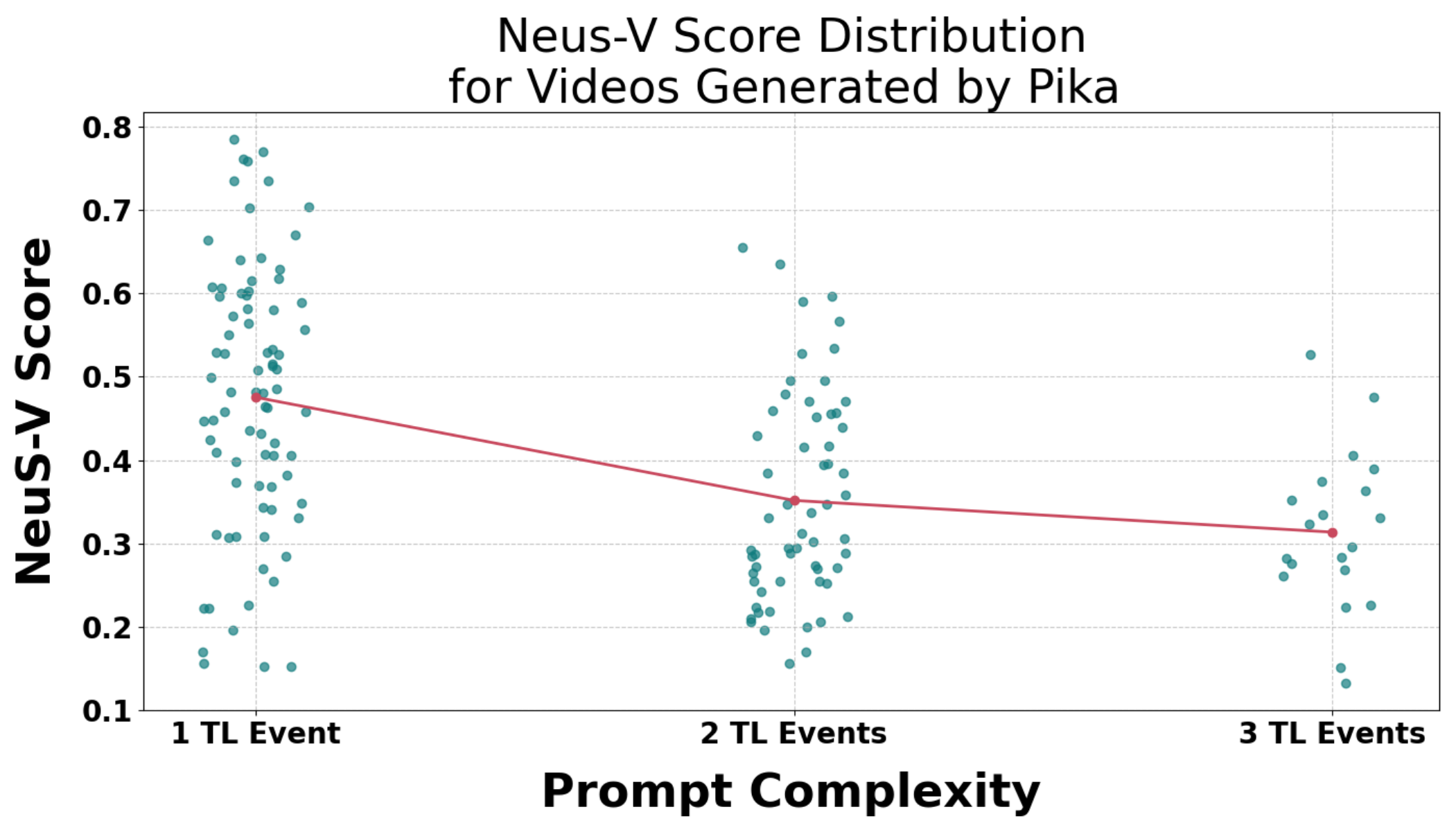}
        }
    }
\end{figure}

\paragraph{Traditional deep learning models struggle to capture complex and long-term temporal dependencies in videos.} We attribute this to the temporal aggregation of semantic and activity-related deep learning, which couples spatial and temporal feature processing. This limitation is demonstrated with Neuro-Symbolic Video Search with Temporal Logic (NSVS-TL) \citep{choi2024towards}. NSVS-TL, as illustrated by its pipeline in Figure \ref{fig:nsvssample}, maps videos into a probabilistic automaton by converting frames into states using an off-the-shelf neural perception model. Complex user queries are then converted into TL specifications. Consequently, NSVS-TL converts the video search problem into a verification problem and extracts the relevant clips that satisfy the TL specifications corresponding to the user query. By decoupling spatial and temporal understanding, this approach significantly outperforms multimodal foundation models when faced with elaborate user queries, as shown in Figure \ref{fig:nsvstl}.

\paragraph{Foundation models for video generation deteriorate in performance with increasing complexity of the text prompts.} Dozens of text-to-video generation models exist (see Section \ref{sec:rel_vid_gen} for a summary), however, evaluations on benchmarks such as NeuS-V \citep{sharan2024neuro}, with its pipeline depicted in Figure \ref{fig:neusvsample}, demonstrate a deficiency in temporal fidelity, where the semantic alignment between video and prompt deteriorates with increasing user query complexity. Inspired by NSVS-TL, NeuS-V first translates the text prompt into atomic propositions and a TL specification using an optimized large language model (LLM). It then constructs a video automaton representation by assigning semantic confidence scores to these propositions using a vision-language model (VLM). Finally, NeuS-V computes the satisfaction probability by formally verifying the video automaton against the TL specification to produce the final score. In Figure \ref{fig:neusv}, we show that text-to-video generation degrades when the number of events in the query increases. To mitigate this issue, we posit that generative models need to be co-designed with temporal understanding frameworks.

Overall, deep learning models still struggle with both long-form video understanding and video generation. Hence, video agents would need to be built via neuro-symbolic methods that blend deep learning methods with formal logic representations and state machines.

\section{Related Work}

\paragraph{Video Search}
Existing research in video event detection predominantly focuses on tracking spatio-temporal object information using deep neural networks to learn latent representations, such as motion and position, for event detection and classification \citep{event-detect-spatiotemporal, event-detection-video-stream, lstm-event-detect, cnn-event-detect, zheng2022abnormal, interpretable-anomaly-detect, action-detect, action-detect-2}. These methods, while effective, demand substantial computational resources for training and inference \citep{lstm-event-detect, action-detect, action-detect-2}. Some works also extend this to natural language-based event detection, employing VLMs like Video-LLaMA \citep{damonlpsg2023videollama} and Video-ChatGPT \citep{maaz2023video}, which integrate language foundation models such as GPT-4 \citep{openai2023gpt4} and LLaMA \citep{touvron2023llama}, for zero-shot event recognition and visual question answering (VQA). However, their reliance on temporal aggregation limits precise frame identification in long videos. Neuro-symbolic methods have been proposed to address this issue by enabling structured reasoning over longer videos when searching for video clips \citep{temporal-event-detect, yang2023specification, choi2024towards, choi2025real}.

\paragraph{Video Agents}
Recent advancements in video agents have leveraged LLMs and VLMs to enable decision-making, such as ReACT \citep{yao2023react} powered through Langchain \citep{langchain} and ToolLLM \citep{qin2023toolllm}. These have been applied in the video domain where tools have been integrated with VLMs for tasks such as long-from video understanding \citep{jeoung2024adaptive, wang2024videoagent}, video generation for robotics \citep{soni2024videoagent}, and video editing \citep{wang2024lave}. While these agents demonstrate progress, they cannot be applied in real-world applications, such as home security, which require the seamless integration of temporal reasoning, context-aware decision-making, and robust interaction with external systems.

\paragraph{Video Generation}
\label{sec:rel_vid_gen}
Following the recent successes in text-to-image generation, text-to-video models such as SORA from OpenAI \citep{openai2024sora}, GEN-3 Alpha from Runway \citep{runway2024gen3}, and PIKA \citep{pikaai2024} have seen increased proliferation. Although the exact architecture for these models is unknown, text-to-video generation employs diffusion models \citep{ho2020denoising}, such as in
Phenaki \citep{villegas2022phenaki} and I2VGen-XL \citep{ zhang2023i2vgen, blattmann2023align, esser2023structure}, or autoregressive models, such as the CogVideo series \citep{hong2022cogvideo, yang2024cogvideox}. For a detailed survey on these methods, we refer you to \cite{cho2024sora}. These models are inadequate for generating temporally correlated events \citep{sharan2024neuro, choi2025we}, and to our knowledge, neuro-symbolic methods for solving this problem have not been explored.

\paragraph{Neuro-symbolic Methods}
Many works explore approaches to building symbolic representations for video classification \citep{short1, short2}, event detection \citep{event-detection-video-stream, cnn-event-detect, lstm-event-detect}, video question-answering \citep{neural-symbolic, chen2022comphy}, robotics \citep{shoukry2017linear,hasanbeig2019reinforcement,kress2009temporal}, and autonomous driving \citep{jha2018safe,mehdipour2023formal}. These methods either construct graph structures \citep{graphical-model-relationship-detection, visual-symbolic, long3}, use latent-space representations as symbolic representations \citep{symbolic-high-speed-video, BertasiusWT21, neural-symbolic-cv}, or leverage formal language methods \citep{Baier2008} to design specifications.

\section{The Formal Challenge}
Our goal is to encourage the research community to develop the analogous version of the LLM-based agent datasets, tasks, and evaluation frameworks (\eg, Tool Bench \citep{qin2023toolllm}, Stable Tool Bench \citep{guo2024stabletoolbench}, and Gorilla \citep{patil2023gorilla}) for the video domain to create and evaluate video agents. To this end, we formalize the design goals of a video agent that leverages deep-learning and neuro-symbolic methods to process videos and complex natural language queries. The inputs and outputs of such a system are:

\begin{itemize}[itemsep=0.5pt, topsep=0pt]
    \item \textbf{Inputs}
        \begin{itemize} [itemsep=0.5pt, topsep=0pt]
            \item \textbf{Dataset:} Videos annotated with natural language queries for temporally complex events and their corresponding ground truth spans or actions. We provide a preliminary dataset used for video search, the TLV Dataset in Section \ref{sec:datasets}, for the community to build upon and adapt for video agents.
            \item \textbf{Tools}: A set of apps or programs to be executed, such as calling Python code, executing a state machine, or calling an API from an external library like Twilio \citep{twilio} and RapidAPI \citep{rapidapi}. We encourage the community to develop open-sourced API's specific to video agent use-cases. 
            \item \textbf{Models:} Deep learning models such as LLMs, VLMs, and Generative Models.
        \end{itemize}
    \item \textbf{Outputs}
        \begin{itemize} [itemsep=0.5pt, topsep=0pt]
            \item \textbf{Event Clips:} A sequence of specific frames that correspond directly to a user query.
            \item \textbf{Actions:} A tool invoked with its inputs timed according to the user query.
            \item \textbf{Synthetic Videos:} Synthesized videos based on user queries and specifications.
        \end{itemize}

    \item \textbf{Metrics}
            \begin{itemize} [itemsep=0.5pt, topsep=0pt]
            \item \textbf{Accuracy of Events:} F1-Score between the predicted spans and ground-truth spans. 
            \item \textbf{Tool Calling:} Accuracy of the selected tool and its desired input  (see Section \ref{sec:metrics}). 
            \item \textbf{Synthetic Videos:} VBench \citep{huang2024vbench} for visual quality and NeuS-V \citep{sharan2024neuro} for temporal fidelity.
        \end{itemize}
\end{itemize}

The primary objective of the video agent is to merge agentic workflows capable of tool invocation \citep{langchain} based on input videos and queries. Therefore, given a dataset annotated with events and corresponding tool actions, the agent would predict which tool should be invoked for each query at specific instances in the video. The core technical requirements for each of the capabilities for the video agent are:

\begin{enumerate}
    \item \textbf{Video Search:} 
    The video search task aims to predict the temporal span of a video clip that corresponds to a given natural language query.  
    This requires:
        \begin{enumerate}[itemsep=0.5pt, topsep=0pt]
            \item \textbf{Parsing Queries:}
            Decompose the natural language query into semantic components that include objects and short-term events or activities followed by a formal language description of their temporal order.
            \item \textbf{Perception:} Orchestrate neural models to detect relevant objects and activities.
            \item \textbf{Prediction:} Predict temporal spans in the query-aligned video with high probability.
        \end{enumerate}
    \item \textbf{Tool Calling:} 
    The key capability of the video agent is to be able to call the right tools and specify their input based on the user's specifications and the context developed from the video.
    The implementation of this task requires:
        \begin{enumerate}[itemsep=0.5pt, topsep=0pt]
        \item \textbf{Tool Selection}: Run deep learning models to identify the right tool or API to call based on the identified temporal spans from the video search task.
    
        \item \textbf{Tool Invocation:} Specify the inputs to the tool to be executed based on the identified video clip---for instance, the message to be sent for a notification API. 
    \end{enumerate}

    \item \textbf{Video Generation:} The video generation task aims to synthesize videos that align with a temporally extended natural language query. To accomplish this, the following are necessary:

    \begin{enumerate}[itemsep=0.5pt, topsep=0pt]
        \item \textbf{Synthesis:} Synthesize novel scenarios based on the user's custom queries.
        \item \textbf{Evaluation:} Ensure high visual quality while maintaining semantic coherence and accurate temporal alignment with the query.
        \item \textbf{Improvement and Editing:} Iteratively improve the videos through reprompting or editing from neuro-symbolic feedback to meet the user's specifications.
    \end{enumerate}
\end{enumerate}

We provide initial data along with benchmarks and example agents for the community to build upon on \href{https://github.com/UTAustin-SwarmLab/Neuro-Symbolic-Agent-Challenge}{GitHub}.

\section{Datasets}
\begin{figure}[t]
\captionsetup{type=figure}
    \includegraphics[width=\linewidth]{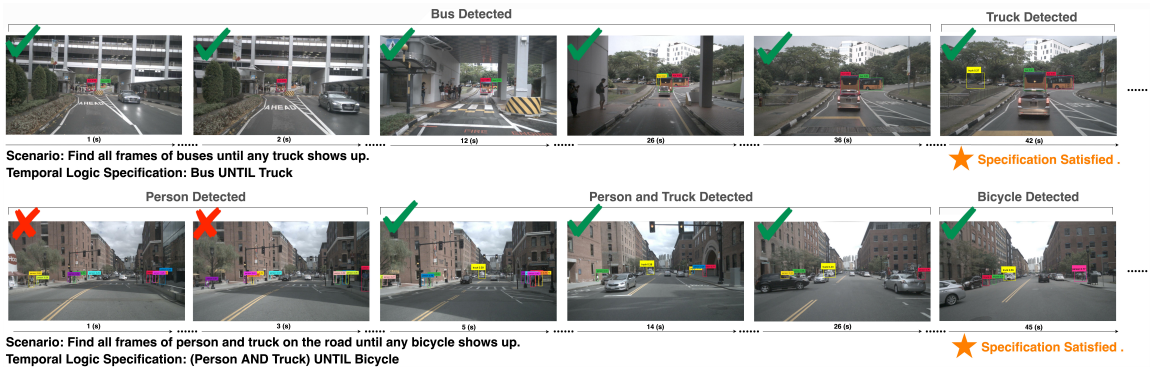}
    \captionof{figure}{\textbf{TLV Dataset Specification-Video Pairing.} An excerpt of the TLV dataset is shown here, demonstrating the efficacy of the TLV dataset in showing the relationships between videos and TL sequences. This figure was taken with permission from \cite{choi2024towards}.}
    \label{fig:tlv}
\end{figure}

\label{sec:datasets}

A plethora of video datasets, such as Ego4D \citep{grauman2022ego4d}, MSR-VTT \citep{xu2016msr}, and others \citep{nagrani2022learning, chandrasegaran2025hourvideo, wang2023internvid}, annotate short video clips with activities in natural language. However, \textbf{they are not suitable for our purpose} because they lack annotations for temporally structured activities such as ``The garage door opens after the person is identified as a delivery driver.''

To develop video agent capabilities such as search, generation, and tool invocation, datasets must include three key annotations: (1) frame-level temporal annotations for events (\eg, detecting deliveries or break-ins), (2) specifications of tools and their inputs (\eg, notification type and recipient), and (3) the temporal order of tool invocations (\eg, ``alert homeowner before locking doors"). 

\paragraph{The TLV dataset is a first step towards addressing these limitations.} The TLV dataset \citep{choi2024towards} recognizes the need to explicitly annotate \textit{when} an event occurs and how they are \textit{temporally related}. At a high level, the TLV dataset is designed with frame-level temporal annotations for temporally dependent events. These annotations are compiled from a combination of static images from leading image datasets, including Waymo \citep{sun2020scalability} and NuScenes \citep{nuscenes}, as illustrated in Figure \ref{fig:tlv}, and the dataset is publicly accessible through \href{https://huggingface.co/datasets/minkyuchoi/Temporal-Logic-Video-Dataset}{Hugging Face}. While TLV facilitates evaluations on video content for video search with temporally correlated event queries, it lacks the annotations pertaining to the tool invocation. This poses an interesting challenge, and we empower the community to curate temporally meaningful datasets with timestamped annotations of the tools used on the more modern video-activity datasets listed above.

\section{Metrics}
\label{sec:metrics}
Consider our running example described in Section \ref{sec:intro}.
\begin{quote}\textit{How can we evaluate the success of a video agent that is capable of video search and toolchain execution in both real-world and good quality synthetically generated scenarios?}
\end{quote}

A comprehensive evaluation metric for agents would consider computational efficiency, video processing latency, state machine generation latency, and resource consumption, such as energy and memory. However, for this challenge, we will focus specifically on evaluating the accuracy of the video agent and video generation.

\paragraph{Event Search Accuracy for Video Search} In the video search domain, we propose utilizing classification metrics, such as the F1 score. This metric, the aggregation of precision and recall, is calculated based on the accurate identification of frames relevant to a given search query compared against ground truth labels provided by datasets like the TLV dataset.

\begin{figure}[t]
\captionsetup{type=figure}
    \centering
    \includegraphics[width=0.98\linewidth]{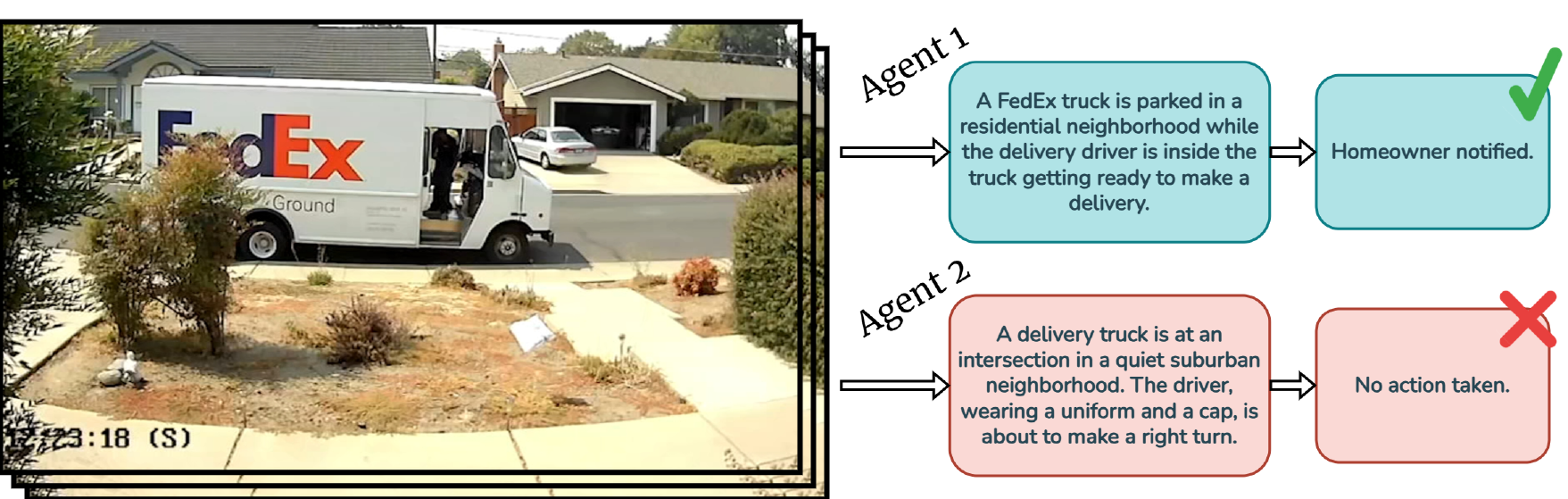}
    \captionof{figure}{\textbf{What is the Correct Action for the Agent?} At a first glance, both agents summarize the video nearly identically. However, upon closer inspection, Agent 1, although more vague, correctly identifies the parked delivery truck and notifies the homeowner. In contrast, Agent 2, while more detailed, hallucinates a right turn on an intersection, leading to no action.}
    \label{fig:fedex}
\end{figure}

\paragraph{Evaluating Tool Calling} 

A comprehensive metric to evaluate video agents would ensure that agents interpret events accurately, maintain temporal coherence with user specifications, and execute actions with appropriate inputs. We list the components as follows:

\begin{itemize}
    \item \textbf{Event-Specific Tool Invocation:} The metric must assess the accuracy of the tools executed in response to specific events. For instance, in Figure \ref{fig:fedex}, the agent must correctly notify the homeowner when a delivery person is detected approaching.
    
    \item \textbf{Temporal Alignment with the Prompt:} The evaluation must ensure that the temporal order of tool invocation adheres to the specified sequence. For example, the garage door should be closed only after the delivery person leaves the driveway but before they leave the package. Similarly, if a break-in is detected, the system must simultaneously trigger an alarm, lock all entry points, and notify law enforcement authorities.

    \item \textbf{Correct Inputs to Tools:} The tools invoked by the agent must receive accurate and contextually appropriate inputs. For instance, notifications should be sent to the correct recipient (\eg, notifying law enforcement in case of a break-in).
\end{itemize}

\paragraph{Video Generation Evaluation} Video generation necessitates an assessment of both visual quality and temporal fidelity. Existing evaluation metrics, such as those based solely on VQA \citep{wu2023discovqa, wu2023exploring}, emphasize visual quality, neglecting temporal coherence in the process. Consequently, these metrics may yield high scores, even for videos that fail to represent the prompt's intended sequence of events. To address this deficiency, we propose a two-pronged evaluation approach: one benchmark to assess visual quality and another to evaluate temporal adherence. Specifically, we suggest employing VBench \citep{huang2024vbench} as the standard for visual quality assessment, complemented by NeuS-V \citep{sharan2024neuro}, a metric we have discussed earlier that conducts temporal coherence evaluations on the generated video by converting its natural language prompt into atomic events and semantics with TL specifications. 

\section{Architecture Discussion and Open Questions}
\label{sec:arch_disc}

In this section, we pose several questions and ideas to the research community to expand the scope for neuro-symbolic video agents to multimodal inputs, multiple formal language representations, and multiagent setups.

\paragraph{Multimodal: Audio, Video, and Beyond.} 
At its core, our description of a video agent decomposes user queries into fundamental atomic events that can be detected using computer vision. However, consider the case where a video agent must respond to multimodal triggers, such as a home security system awaiting the audio of a door knock. How can I modify my agent framework to allow for complex, multimodal triggers? For this system, the role of the neural model extends beyond pure video analysis, encompassing the complex orchestration of multimodal information to facilitate robust and adaptable symbolic reasoning.

\paragraph{How is Formal Language Specified?}
The temporal structure of events through formal language presents a unique challenge. From Section \ref{sec:intro}, we described formal language through TL, allowing the video agent to capture temporal information in user queries. However, video agents are not married to a single type of formal language framework; if precise timings were vital, video agents can leverage approaches such as Linear Temporal Logic (LTL) or Signal Temporal Logic (STL). Now, consider the need to track spatial relationships or distances between objects. This idea warrants the exploration of more flexible formal language, such as state machines or pipelines akin to LangChain \citep{langchain}. Ultimately, any approach must be both expressive and scalable, capable of capturing complex temporal dependencies while remaining computationally feasible for verification. Further exploration is encouraged to determine which approach, or combination of approaches, yields the most robust representation framework.

\paragraph{Multi-Agent Equals Multi-Camera?}
Consider an agentic security system developed for a large apartment complex with dozens of security cameras instead of one. How can these cameras interface with each other to create one cohesive system? This now becomes a multi-agent problem where atomic propositions can be derived from different security cameras across the complex. This introduces several key challenges: What mechanisms facilitate the fusion of disparate spatial and temporal information into a unified representation? Furthermore, how can the system reason about events occurring across multiple viewpoints, effectively resolving potential ambiguities and inconsistencies? The goal for a multi-agent system is to move beyond the simple aggregation of camera data and towards a truly robust system that can reason about complex, distributed events with enhanced situational awareness.

\section{Conclusion}
The development of neuro-symbolic video agents mark the next frontier in video understanding, blending the pattern recognition capabilities of deep learning with the interpretability and temporal reasoning strengths of symbolic methods. These hybrid systems address critical limitations of current models, such as poor long-term memory and a lack of guarantees for perception and action. By bridging low-level recognition with high-level logic, neuro-symbolic approaches enable robust performance in complex settings like home security, autonomous driving, and beyond. They empower systems to  reason over temporal sequences, generate synthetic content for edge cases, and seamlessly translate understanding into action. We encourage the research community to expand this paradigm by exploring architectures and methods that tightly integrate learning with formal reasoning, paving the way toward truly intelligent video agents.

\acks{This material is based upon work supported in part by the Office of Naval Research (ONR) under Grant No. N00014-22-1-2254. Additionally, this work was supported by the Defense Advanced Research Projects Agency (DARPA) contract DARPA ANSR: RTX CW2231110. Approved for Public Release, Distribution Unlimited.}

\bibliography{references}

\end{document}